\documentclass[10pt,twocolumn,letterpaper]{article}

\usepackage[pagenumbers]{cvpr} 

\usepackage{graphicx}
\usepackage{amsmath}
\usepackage{amssymb}
\usepackage{booktabs}
\newlength\savewidth\newcommand\shline{\noalign{\global\savewidth\arrayrulewidth
  \global\arrayrulewidth 1pt}\hline\noalign{\global\arrayrulewidth\savewidth}}
\newcommand{\tablestyle}[2]{\setlength{\tabcolsep}{#1}\renewcommand{\arraystretch}{#2}\centering\footnotesize}
\usepackage{color}
\definecolor{baselinecolor}{gray}{.92}

\usepackage[pagebackref,breaklinks,colorlinks]{hyperref}

\usepackage[capitalize]{cleveref}
\crefname{section}{Sec.}{Secs.}
\Crefname{section}{Section}{Sections}
\Crefname{table}{Table}{Tables}
\crefname{table}{Tab.}{Tabs.}

\def\cvprPaperID{4494} 
\def\confName{CVPR}
\def\confYear{2023}

\begin{document}

\title{Self-supervised Learning by View Synthesis}

\author{%
  Shaoteng Liu\textsuperscript{$1$} \qquad
  Xiangyu Zhang\textsuperscript{$2$} \qquad
  Tao Hu\textsuperscript{$1$} \qquad
  Jiaya Jia\textsuperscript{$1,3$} \\
  \textsuperscript{$1$}The Chinese University of Hong Kong \quad
  \textsuperscript{$2$}MEGVII Technology \quad
  \textsuperscript{$3$}SmartMore \\
  \tt\small \{stliu21,taohu,leojia\}@cse.cuhk.edu.hk \qquad
  \tt zhangxiangyu@megvii.com \\
}
\maketitle

\begin{abstract}
We present view-synthesis autoencoders (VSA) in this paper, which is a self-supervised learning framework designed for vision transformers. Different from traditional 2D pre-training methods, VSA can be pre-trained with multi-view data. In each iteration, the input to VSA is one view (or multiple views) of a 3D object and the output is a synthesized image in another target pose. The decoder of VSA has several cross-attention blocks, which use the source view as value, source pose as key, and target pose as query. They achieve cross-attention to synthesize the target view. This simple approach realizes large-angle view synthesis and learns spatial invariant representation, where the latter is decent initialization for transformers on downstream tasks, such as 3D classification on ModelNet40, ShapeNet Core55, and ScanObjectNN. VSA outperforms existing methods significantly for linear probing and is competitive for fine-tuning.
The code will be made publicly available.
\end{abstract}

\section{Introduction}
\begin{figure}[t]
\centering
\includegraphics[width=1.0\linewidth]{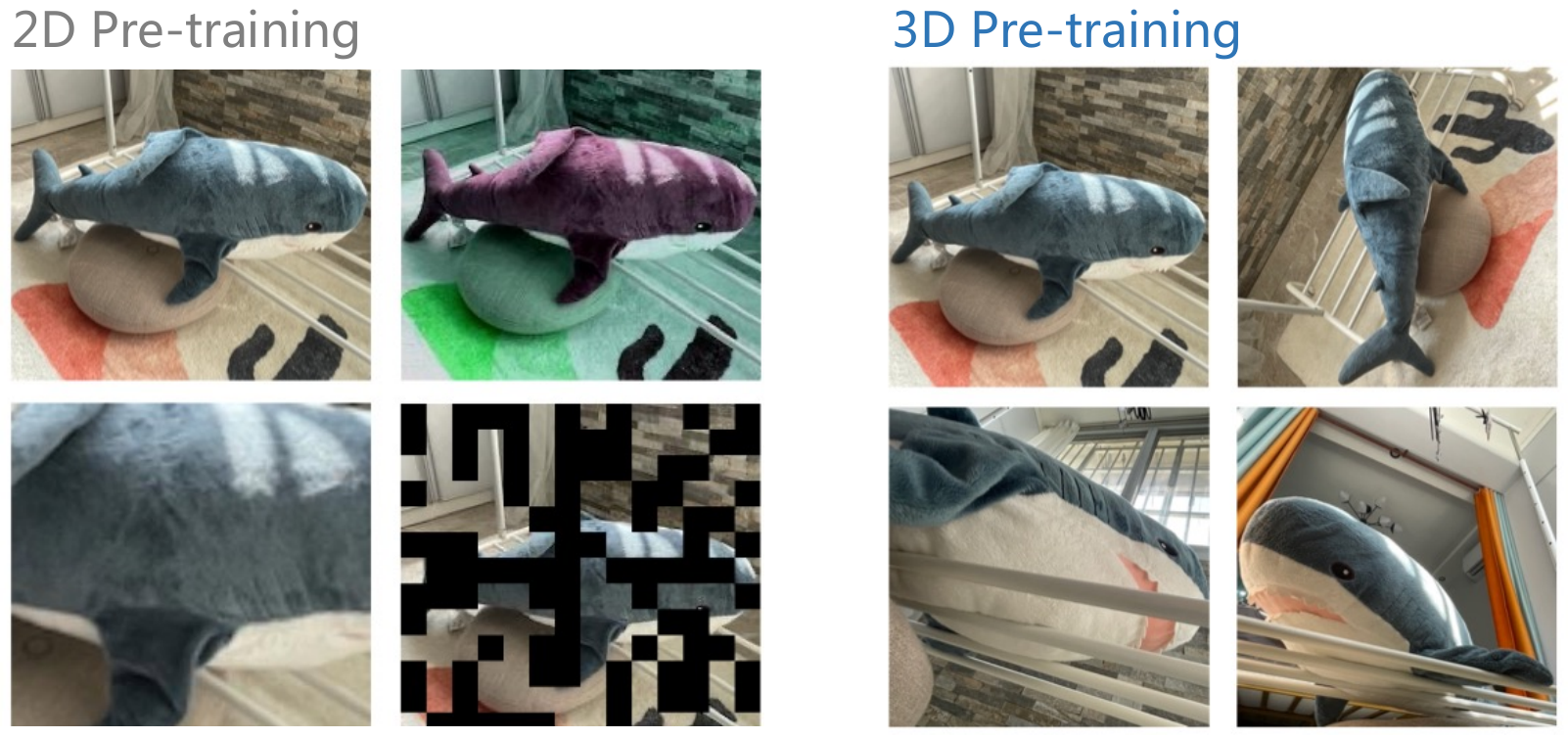}
\caption{Left: Traditional 2D self-supervised learning methods learn with masks or augmentations (crop, colorization, ...). Right: Our 3D self-supervised learning method learns with multi-view data, yielding a new task.}
\label{fig:tissue}
\end{figure}

Self-supervised learning makes it possible to train deep learning models with infinite unlabeled data~\cite{he2020momentum,he2021masked}. The unlabeled data has different structures~\cite{reizenstein2021common} and is general to learn semantic information. It is notable that, so far, there is still no much work to consider inherent 3D geometry connections among input data. An example is shown in Fig.~\ref{fig:tissue} with multi-view shark images. This multi-view setting is not considered yet in most previous self-supervised learning work. Existing methods treat them, among millions of images, as different instances. 
In this paper, we make the first attempt to pre-train vision transformers with multi-view unlabeled images. 

Although CNNs~\cite{krizhevsky2012imagenet} are still the most popular vision models, previous study shows that these grid-wise methods cannot model robust representation for 3D data~\cite{hinton2021represent}.
Vision transformers are more suitable and flexible for this problem.
Pre-training, on the other hand, is important for vision transformers~\cite{bao2021beit,he2021masked}, since it is related to inductive bias. Unlike CNNs, transformers such as ViT~\cite{dosovitskiy2020image} yield less inductive bias for vision in the model structure. Several recent methods have proved that less inductive bias may result in larger capacity of vision models~\cite{dosovitskiy2020image,liu2021swin}, making them perform better in longer training schedule. Good initialization by pre-training gives these larger capacity models correct direction to learn~\cite{dosovitskiy2020image,zheng2021rethinking}.
This is vital for the whole system.

Pre-training for vision transformers was extensively studied, such as DINO~\cite{caron2021emerging}, BEiT~\cite{bao2021beit} and MAE~\cite{he2021masked}. It is noteworthy that the self-supervised methods in fact learn something invariant. For example, MAE is to learn mask-invariance where the semantic of an image is mostly invariant with patches masked. Contrastive learning enforces augmentation-invariance where the semantic of an image is invariant after rotation, flipping, cropping and colorization. Several examples are shown on the left side of Fig.~\ref{fig:tissue}. 

In addition to above image-based invariance, in this paper, we advocate important 3D spatial-invariance and study its applicability and generality. On the right-hand side of Fig.~\ref{fig:tissue}, the shark toy is observed from different views. They refer to the same instance, albeit apparently different in terms of image content. This type of information was not taken into consideration in self-supervised learning. We in this paper handle ubiquitous multi-view data, utilizing the new spatial invariance constraint.

\begin{figure*}[t]
\centering
\includegraphics[width=1\linewidth]{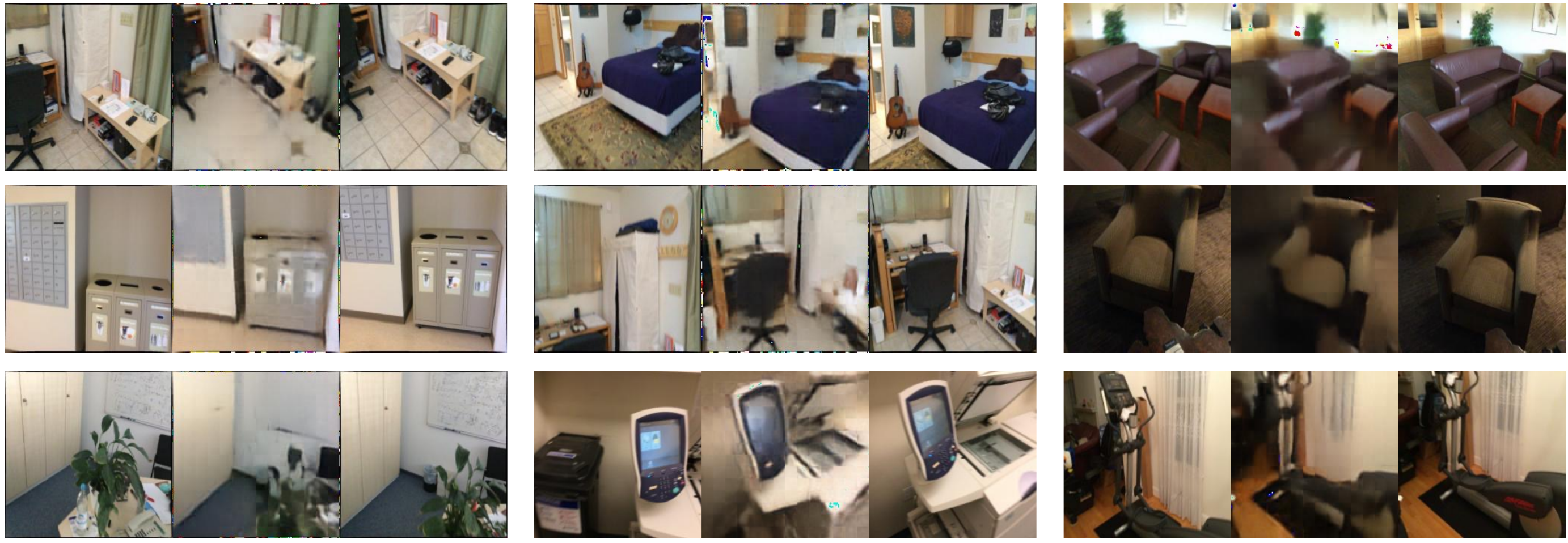}
\caption{Exemplar results on Scannet~\cite{dai2017scannet}. For each triplet, we show source image (left), image synthesized by VSA (middle) and target image (right). A simple transformer is already able to synthesize high-quality images.}
\label{fig:scannet}
\end{figure*}

\begin{figure*}[t]
\centering
\includegraphics[width=1.0\linewidth]{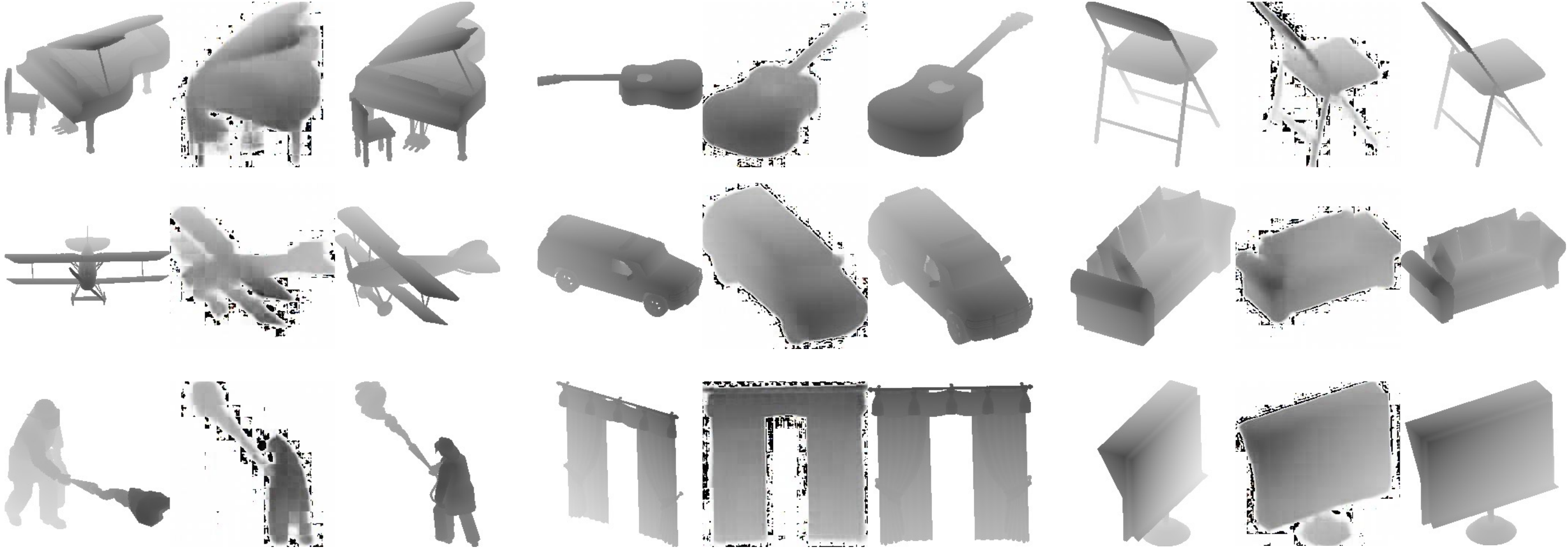}
\caption{\textbf{Exemplar results on ModelNet40.} For each triplet, we show source image (left), image synthesized by VSA (middle) and target image (right).}
\label{fig:modelnet}
\end{figure*}

To this end, we propose view-synthesis autoencoders (VSA), a self-supervised learning framework for vision transformers.
VSA is trained with multi-view data, instead of traditional i.i.d images. In each iteration, VSA takes one view (or multiple views) of an object as the source image, and aims to synthesize a target view. The decoder of VSA has several cross-attention blocks, which use the source view as value, source pose as key, and target pose as query. VSA accomplishes large-angle view synthesis as shown in Figs.~\ref{fig:scannet}, ~\ref{fig:modelnet}, and \ref{fig:modelnet_mask}. By completing synthesis, it also learns spatial invariant representation. As validated in our experiments, the representation benefits downstream tasks such as 3D classification on ModelNet40 \cite{wu20153d}, ShapeNet Core55 \cite{chang2015shapenet} and on ScanObjectNN \cite{uy2019revisiting}.
The total contribution is the following.
\begin{itemize}
    \item \emph{Novel and important task}: As far as we know, VSA is the first framework for self-supervised learning with transformers using multi-view data. It changes the original i.i.d training pipeline.
    \item \emph{General solution}: Any feature encoder can be directly moved into VSA. It is natural to achieve view synthesis in our cross-attention structure.
    \item \emph{Competitive accuracy}: Without any more complicated encoder design and view sampling strategy, VSA achieves competitive accuracy on 3D classification benchmarks. It outperforms existing pre-training methods notably for linear probing.
\end{itemize}

\section{Related Work}

\noindent\textbf{Vision Transformer.}
Recently, vision transformers show the great power to substitute convNets  (CNNs), achieving comparable or even better results \cite{krizhevsky2012imagenet}. 
iGPT~\cite{chen2020generative} applies transformer to image pixels by reducing image resolution.
ViT~\cite{dosovitskiy2020image} is the transformer to classify images with image patches. Besides, vision transformers have been applied to various tasks, including semantic segmentation~\cite{zheng2021rethinking,xie2021segformer,cheng2021masked}, object detection~\cite{carion2020end,zhu2020deformable} and image processing~\cite{chen2021pre}. A lot of ViT variants~\cite{touvron2021training,wang2021pyramid,han2021transformer,liu2021swin} appear and benefit applications.

\noindent\textbf{Self-supervised Learning.}
Self-supervised learning becomes popular. Various pre-training tasks have been proposed. Contrastive learning forms the mainstream methods~\cite{becker1992self,hadsell2006dimensionality} before. For example, InsDis~\cite{wu2018unsupervised}, CMC~\cite{tian2020contrastive}, MoCo~\cite{he2020momentum} and SimCLR~\cite{chen2020simple} find different ways to create positive and negative pairs. These pairs can help model similarity and dissimilarity between views. Besides, in BYOL~\cite{grill2020bootstrap} and SimSiam~\cite{chen2021exploring}, authors point out that negative pairs may not be necessary. Contrastive learning has also been introduced to transformer pre-training, such as DINO~\cite{caron2021emerging} and MoCo v3~\cite{chen2021empirical}. All of these methods depend heavily on the design of augmentation.

Other than contrastive learning, masked image modeling attracts attention recently. BERT~\cite{devlin2018bert} and GPT~\cite{radford2018improving,radford2019language,brown2020language} are successful for NLP pre-training. iGPT~\cite{chen2020generative} follows GPT to predict unkown image pixels. BEiT~\cite{bao2021beit} predicts discrete tokens, like~\cite{van2017neural}. MAE~\cite{he2021masked} reconstructs images given mask-out. A lot of followups appear these days~\cite{zhou2021ibot,xie2021simmim,wei2021masked,chen2022context}.
They prove to be suitable for vision transformer pre-training. These methods learn augmentation- and mask-invariance.

\noindent\textbf{3D Pretraining.}
There is work to explore the way to learn representations from 3D priors. Pri3D\cite{hou2021pri3d} employs contrastive learning with multi-view images and point-clouds. It uses corresponding pixels in different views as a positive pair. It proves that 3D priors can benefit 2D downstream tasks. Pri3D is designed for CNNs. Instead of performing point-level contrastive learning, our VSA directly synthesizes a whole image based on another view. This process is more natural for vision transformers and can be simply achieved by a cross-attention block.

\noindent\textbf{Multi-View 3D Shape Classification.}
It is first proposed in~\cite{bradski1994recognition} to recognize 3D objects using multiple 2D images. MVCNN~\cite{su2015multi} introduces deep CNNs and view pooling in this task. Followups modify the view fusion strategy to improve performance~\cite{yu2018multi,yang2019learning,feng2018gvcnn}. RotationNet~\cite{kanezaki2018rotationnet} jointly recognizes images and poses. ViewGCN~\cite{wei2020view} uses a dynamic graph CNNs to pool features from different views. Recently, methods select views instead of using fixed views. VERAM~\cite{chen2018veram} selects views by reinforcement learning. MVTN~\cite{hamdi2021mvtn} jointly trains a model with a multi-view task-specific network. In VSA, we only use the fixed-view pipeline because of the need for an ablation study to clearly know effectiveness of our modules. 

\noindent\textbf{Single-view Reconstruction.}
Single-view reconstruction is to reconstruct a 3D shape from a single image~\cite{choy20163d,girdhar2016learning}. Single-view reconstruction methods can be divided into 3D supervised~\cite{wang2018pixel2mesh,gkioxari2019mesh,mescheder2019occupancy,fan2017point}, 2D supervised~\cite{kato2018neural,liu2019soft,chen2019learning} and unsupervised reconstruction~\cite{li2020self,navaneet2020image,insafutdinov2018unsupervised}. Some works also use neural radiance field~\cite{mildenhall2020nerf,rematas2021sharf} for single-image novel view synthesis. Most of these methods~\cite{li2020self,niemeyer2020differentiable} not only reason about the visible 3D structure in a 2D image, but also reconstruct invisible parts by learning prior knowledge from big data. A previous study~\cite{tatarchenko2019single} points out that sometimes doing reconstruction is in fact doing recognition. Since single-view reconstruction is a process to learn high-level knowledge, it is interesting to explore whether it is a way to do self-supervised learning.

\section{Our Approach}

\begin{figure*}[t]
\centering
\includegraphics[width=1\linewidth]{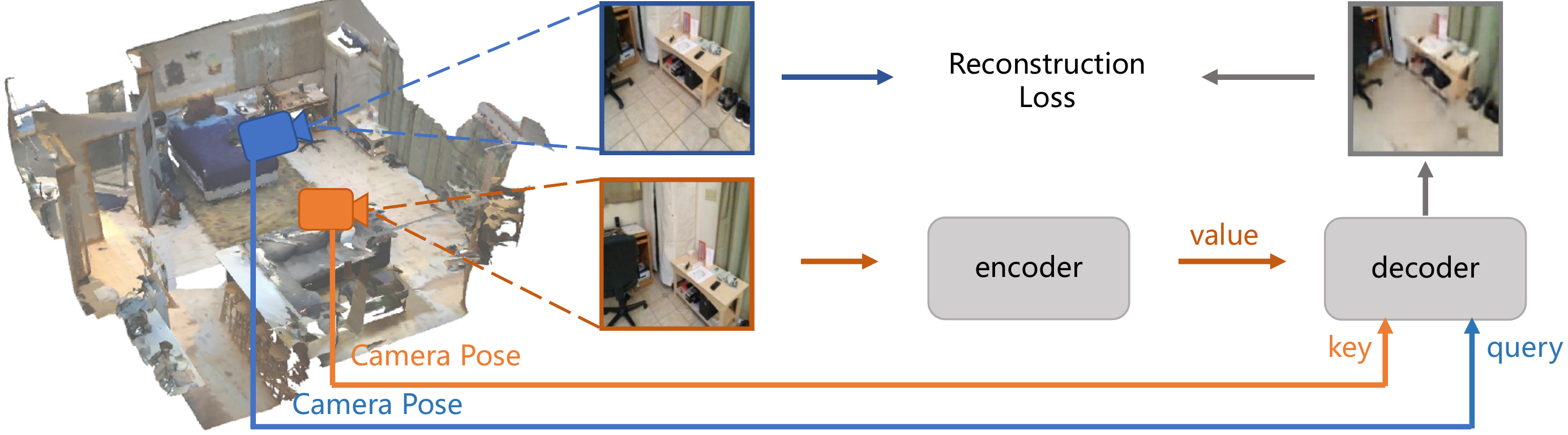}
\caption{\textbf{Framework of VSA.} The orange camera gives the source pose and the blue one yields the target pose. The source view is fed into the encoder (any feature extractor can be used). There are several cross-attention blocks in the decoder. The output of decoder is a synthesized image, which is used to calculate the MSE loss with the target view. This example is from ScanNet~\cite{dai2017scannet}.}
\label{fig:framework}
\end{figure*}

Our view-synthesis autoencoder (VSA) synthesizes a new view of an object based on the input view of it. Unlike other typical autoencoders, VSA is not designed to recover the input signal. Only if the input and query views are the same, it degrades to a common autoencoder. The framework of VSA is shown in Fig.~\ref{fig:framework}.

\begin{figure*}[t]
\centering
\includegraphics[width=0.85\linewidth]{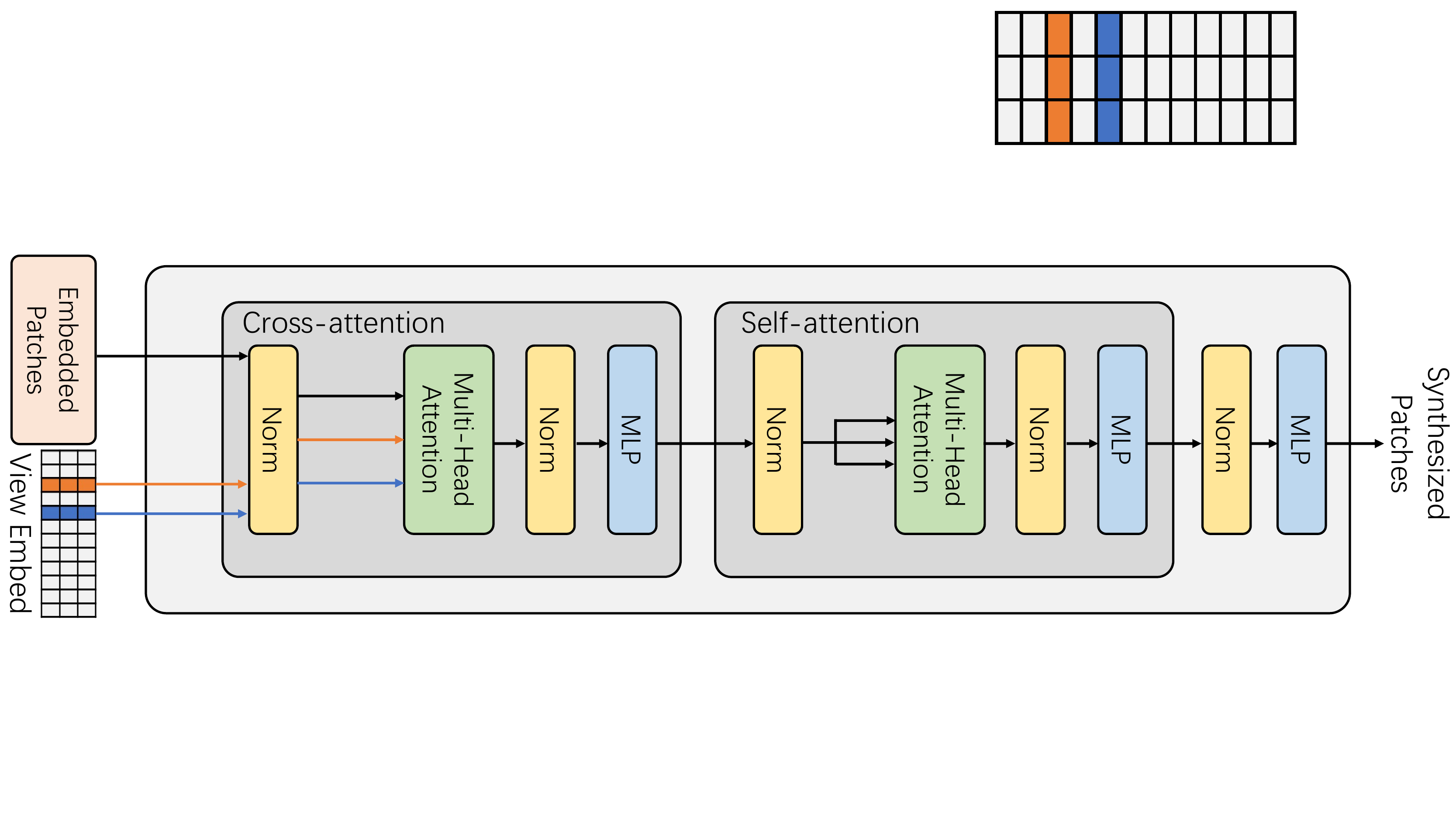}
\caption{\textbf{Architecture of VSA decoder.} Upper: detailed architecture of VSA decoder. Lower: The embedded patches are the output of VSA encoder. View embedding is a learnable matrix. With a single source view (lower left), orange and blue vectors are corresponding to source and target pose. In a cross-attention block, embedded patches are the value (black arrow), source pose is the key (orange arrow), and target pose is the query (blue arrow). With two source views (lower right), two source poses (yellow and orange) and two embedded patches are respectively concatenated before feeding into VSA decoder.}
\label{fig:arch}
\end{figure*}

\noindent\textbf{VSA Encoder.}
The encoder of VSA is simply a ViT~\cite{dosovitskiy2020image}. This part is not our contribution. ViT-B is the default architecture. No special design is needed here. Thus any feature extractors work.
The input to VSA encoder is a view of an object/scene, as shown in Fig.~\ref{fig:framework}.
The output of VSA encoder is the value (red in Fig.~\ref{fig:framework}) for cross-attention blocks in VSA decoder. It is also the Embedded Patches in Fig.~\ref{fig:arch}.

\noindent\textbf{VSA Decoder.}
The architecture of VSA decoder is shown in Fig.~\ref{fig:arch} upper part. Unlike MAE~\cite{he2021masked}, the decoder has cross-attention blocks in the beginning. It takes the value from the output of VSA encoder (black arrow). The key is the embedding of source camera pose (orange arrow) and the query is the embedding of the target camera pose (blue arrow). In Fig.~\ref{fig:framework}, source and target cameras are the orange and blue ones respectively. The process in these cross-attention blocks is the following. To generate a target image, first, we use the corresponding camera pose as query and existing camera pose as key. It computes the attention map across two views to find the parts of information that are useful. Then we extract the feature of the source view as value and synthesize the target view. For the self-attention blocks following the cross-attention blocks, the value, key and query are the same (three black arrows shown in  Fig.~\ref{fig:arch} upper part). The numbers of cross- and self-attention blocks are the same by default. Fig.~\ref{fig:arch} (upper part) simplifies them as one cross-attention block and one self-attention block.

\noindent\textbf{Pose Embedding.}
As described above, both key and query input of cross-attention blocks in VSA decoder are pose embeddings. Here we provide two methods to encode poses. First, if the number of views is fixed as $n$ and the view sampling strategy is constant for all objects in a dataset, we initialize a $n$ dimension learnable parameter. $n=12$ views for ModelNet40 \cite{wu20153d}. The shape of this learnable parameter is ($n$, $patches$, $dim$), where $patches$ is the number of patches for encoder or decoder input, and $dim$ is the dimension of extracted patch features. As shown in Fig.~\ref{fig:arch} lower part, the gray grid is the learnable parameter which has $n$ lines (from left to right, $n=12$ in this example). On the left side of Fig.~\ref{fig:arch} lower part, when the input is a single source view, the orange vector and blue vector are corresponding to the source view and target poses. On the right side of Fig.~\ref{fig:arch} lower part, when the inputs are two source views, the yellow and orange vectors are corresponding to two source poses. The key in cross-attention blocks becomes the concatenation of two source poses embeddings (red). The blue vector is corresponding to the target pose.

Second, if the number of views is varying, such as those in ScanNet \cite{dai2017scannet}, we follow NeRF pipeline~\cite{mildenhall2020nerf} to encode camera poses. Using the camera intrinsic matrix and camera-to-world matrix, we represent a camera view as a bunch of camera rays emitted from the position of the camera. A camera ray can be represented as $r(t)=o+td$, where $o$ is the coordinate of source point and $d$ is the direction of the camera ray. We represent a ray as $concat(o, d)$ in 6D. As a result, the shape of a camera pose embedding is finally ($H$, $W$, 6), which can be directly fed into the VSA decoder without re-shaping.

\noindent\textbf{Multiple Source Views.}
Fig.~\ref{fig:framework} shows the situation of only one source image. In fact, our VSA can take multiple source images as input, as shown in the lower part of Fig.~\ref{fig:arch} (right). Suppose there are two input views, two source images are randomly sampled from all views. They can even be the same. Two source images share the VSA encoder for extracting features. For feature fusion, two features are directly concatenated together, before feeding into the VSA decoder.
Because the patches number of output is only determined by the query, it does not change the output size. Using multiple source views is reasonable, because more information from different angles is provided.
The process in a cross-attention block can be understood as (1) computing the attention map between query positions and existing positions in multiple views, and (2) choosing useful content from multiple views to synthesize a new view.

\noindent\textbf{View Synthesis Loss.}
The view synthesis loss is to compute the mean squared error (MSE) between the synthesized and target images (the images bounded by gray and blue boxes in Fig.~\ref{fig:framework}). The loss is computed in a pixel-by-pixel manner, the same as MAE~\cite{he2021masked}.

\section{Experiments}

We pretrain our proposed VSA on different 3D datasets with multi-view images, including ModelNet40 \cite{wu20153d}, ShapeNet Core55 \cite{chang2015shapenet} and ScanObjectNN \cite{uy2019revisiting}. We apply it to multi-view 3D shape classification. We also show synthesis results on a more realistic dataset of ScanNet \cite{dai2017scannet}.

\begin{table*}[t]
\centering
\subfloat[
\textbf{Decoder depth}. Deeper decoder improves linear probing performance
\label{tab:decoder_depth}
]{
\begin{minipage}{0.3\linewidth}{\begin{center}
\tablestyle{4pt}{1.05}
\begin{tabular}{lcc}
blocks & ft & lin \\
\shline
1 & 91.6 & 70.9 \\
2 & 91.3 & 71.1 \\
4 & \colorbox{baselinecolor}{\textbf{91.7}} & \colorbox{baselinecolor}{\textbf{77.1}} \\
\end{tabular}
\end{center}}\end{minipage}
}
\hfill
\subfloat[
\textbf{Decoder width}. Wider decoder improves linear probing performance
\label{tab:decoder_width}
]{
\begin{minipage}{0.3\linewidth}{\begin{center}
\tablestyle{4pt}{1.05}
\begin{tabular}{lcc}
dim & ft & lin \\
\shline
256 & \textbf{92.0} & 74.8 \\
384 & \colorbox{baselinecolor}{91.7} & \colorbox{baselinecolor}{\textbf{77.1}} \\
512 & 91.9 & 74.6 \\
\end{tabular}
\end{center}}\end{minipage}
}
\hfill
\subfloat[
\textbf{View sampling}. Random sampling works better than fixed sampling
\label{tab:view_sample}
]{
\begin{minipage}{0.3\linewidth}{
\begin{center}
\tablestyle{4pt}{1.05}
\begin{tabular}{lcc}
strategy & ft & lin \\
\shline
random & \colorbox{baselinecolor}{\textbf{91.7}} & \colorbox{baselinecolor}{\textbf{77.1}} \\
fixed & \textbf{91.7} & 70.5 \\
 &  &  \\
\end{tabular}
\end{center}}\end{minipage}
}
\hfill
\subfloat[
\textbf{Cross attention}. Changing all blocks to cross-attention ones hurt performance for fine-tuning
\label{tab:cross_attention}
]{
\begin{minipage}{0.3\linewidth}{\begin{center}
\tablestyle{4pt}{1.05}
\begin{tabular}{lcc}
blocks & ft & lin \\
\shline
1 & 91.6 & \textbf{78.6} \\
2 & \colorbox{baselinecolor}{\textbf{91.7}} & \colorbox{baselinecolor}{77.1} \\
3 & \textbf{91.7} & 75.2 \\
4 & 90.9 & 76.7 \\
\end{tabular}
\end{center}}\end{minipage}
}
\hfill
\subfloat[
\textbf{Source views}. More source views improve linear probing accuracy but affect fine-tuning accuracy
\label{tab:num_views}
]{
\begin{minipage}{0.3\linewidth}{\begin{center}
\tablestyle{4pt}{1.05}
\begin{tabular}{lcc}
case & ft & lin \\
\shline
1 & \colorbox{baselinecolor}{\textbf{91.7}} & \colorbox{baselinecolor}{77.1} \\
2 & 91.1 & \textbf{79.7} \\
4 & 90.8 & 79.5 \\
 &  &  \\
\end{tabular}
\end{center}}\end{minipage}
}
\hfill
\subfloat[
\textbf{Data aug}. rc means random crop. jt means color jitter. They are useful. crop on the target image reduces performance
\label{tab:aug}
]{
\begin{minipage}{0.3\linewidth}{\begin{center}
\tablestyle{4pt}{1.05}
\begin{tabular}{lccc}
src & tar & ft & lin \\
\shline
no & no & 90.2 & 60.4 \\
rc & no & \colorbox{baselinecolor}{\textbf{91.7}} & \colorbox{baselinecolor}{77.1} \\
rc & rc & 90.0 & 73.4 \\
rc, jt & no  & \textbf{91.7} & \textbf{78.6} \\
\end{tabular}
\end{center}}\end{minipage}
}
\caption{\textbf{VSA ablation experiments} with ViT-B on ModelNet40. Fine-tuning (ft) and linear probing (lin) accuracy (\%) are reported. In our default setting, the decoder depth is 4, the decoder width is 384, the source data augmentation is random cropping, no target data augmentation is performed. Masking ratio is 0\%, the number of source views is 1, and the pre-training length is 200 epochs. Default settings are marked in \colorbox{baselinecolor}{gray}. The best results are marked in \textbf{bold}.}
\label{tab:ablations}
\end{table*}

\begin{figure}[t]
\centering
\includegraphics[width=0.9\linewidth]{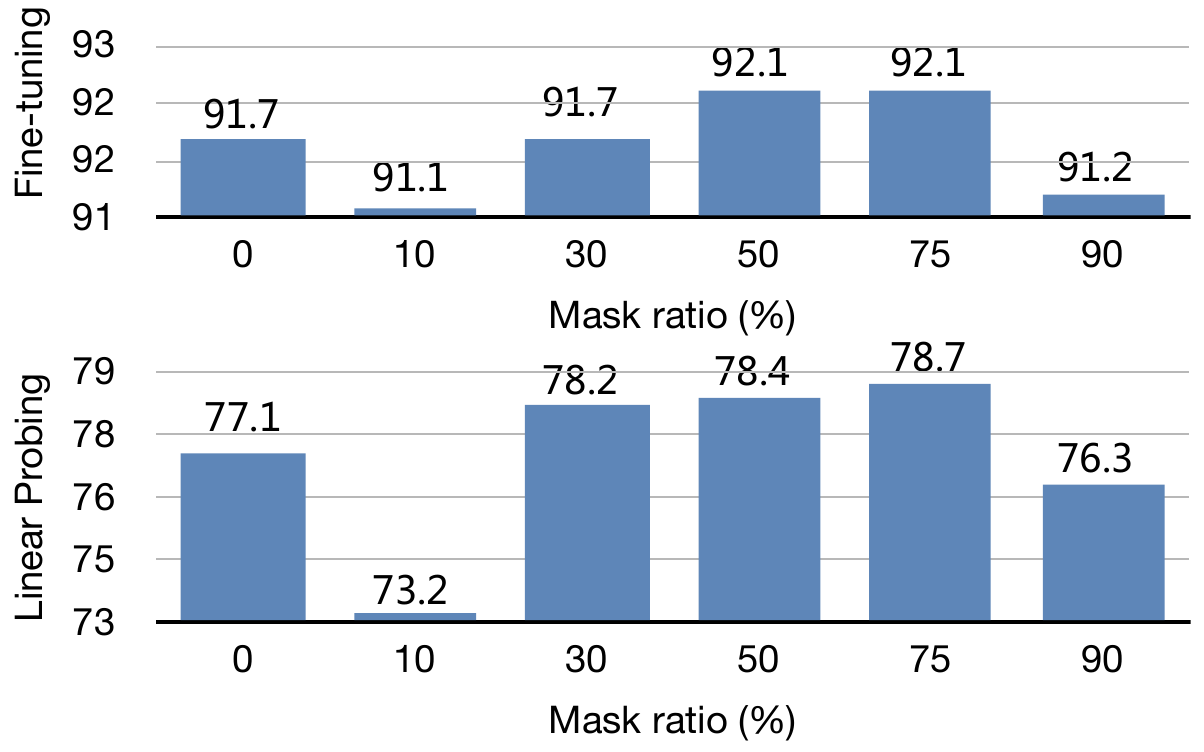}
\caption{\textbf{Masking ratio.} A high making ratio (75\%) can make VSA working better. Masking too many patches (90\%) hurts performance. A small ratio (10\%) is also not recommended.}
\label{fig:mask}
\end{figure}

\begin{figure}[t]
\centering
\includegraphics[width=0.9\linewidth]{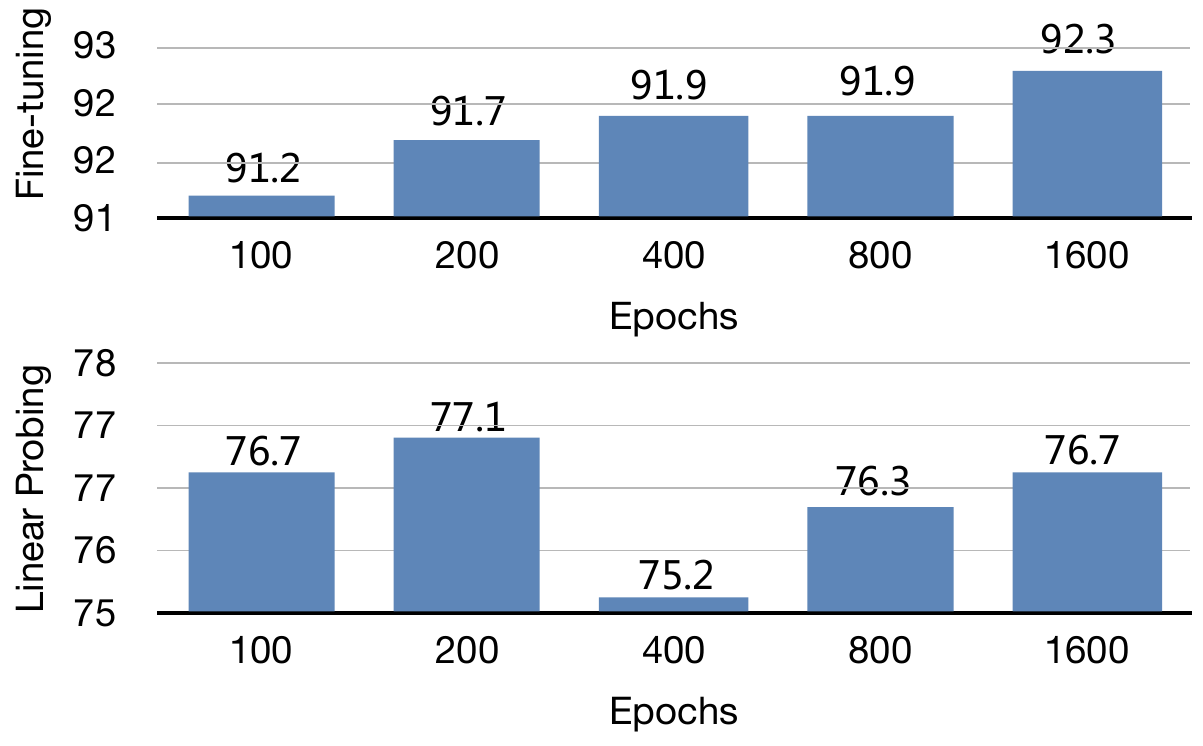}
\caption{\textbf{Training schedules.} A longer training schedule can improve the fine-tuning performance. However, longer training schedule may hurt the linear probing performance.}
\label{fig:epoch}
\end{figure}

\subsection{Datasets}
\noindent\textbf{ModelNet40.} ModelNet40 \cite{wu20153d} consists of 12,311 3D models. A typical dataset splitting strategy is to use 9,843 for training and 2,468 for testing. These models can be labeled with 40 object classes. 2D images are rendered with mesh rendering.
12 views are generated following circular setting in MVTN \cite{hamdi2021mvtn}.

\noindent\textbf{ShapeNet Core55.} ShapeNet Core55 \cite{chang2015shapenet} was proposed for the shape retrieval challenge SHREK \cite{sfikas2017exploiting}.
It contains 51,162 3D mesh objects.
These models can be labeled with 55 object classes. A typical dataset splitting strategy is to use 35,764 for training, 5,133 for validating and 10,265 for testing. 2D images are rendered with point rendering, also in circular setting \cite{hamdi2021mvtn}.

\noindent\textbf{ScanObjectNN.} ScanObjectNN \cite{uy2019revisiting} is a recently proposed point cloud dataset. It is claimed as a more realistic and challenging 3D classification dataset. It has 2,902 models and can be labeled with 15 object classes. MVTN \cite{hamdi2021mvtn} divides it into three main variants: object only, object with background, and the hardest perturbed variant. We only test VSA on the first one. 2D images are rendered with point rendering, also in circular setting \cite{hamdi2021mvtn}.

\noindent\textbf{ScanNet.} ScanNet \cite{dai2017scannet} is an RGB-D video dataset including 2.5 million views from 1,513 ScanNet train video sequences. We first sample every 25 frames from the video sequences. We then filter it, resulting in pairs of frames with $\geq$ 30\% pixel overlap. The pre-training dataset finally has $\approx$840k frame pairs. Following Pri3D \cite{hou2021pri3d}, we obtain pixel overlapping ratio by computing the corresponding geometric chunks of each image.

\subsection{Main Properties}
We do ablation study on ModelNet40 to show the main properties of view-synthesis autoencoders. After pre-training on ModelNet40 training set, we perform both end-to-end fine-tuning and linear probing on the training set. Classification accuracy on ModelNet40 testing set is reported in Table~\ref{tab:ablations}. We use ViT-B as the default encoder.

\noindent\textbf{Decoder Design.}
Decoder is the essential part of VSA. Without using new encoders, our decoder determines new information learned from the pre-training process. Following MAE~\cite{he2021masked}, we ablate the decoder design in Tables~\ref{tab:decoder_depth} and \ref{tab:decoder_width}.
As shown in Table~\ref{tab:decoder_depth}, results line up with MAE. For end-to-end fine-tuning, influence of decoder depth and width vanishes. In contrast, effect enlarges in linear probing. The difference vanishing phenomenon is also observed in MAE. Our experiments manifest that it is not caused by the mask-token task in MAE. Since view-synthesis also causes this effect, it may be caused by the structure of encoder-decoder itself.

In addition, deeper decoder may not always benefit fine-tuning, since we observe that 1-block depth works better than 2-block one. Besides, in Table~\ref{tab:decoder_width}, we observe that the performance of fine-tuning and linear probing may not always be consistent. For example, 256 dim works better in fine-tuning, while 384 dim works better in linear probing.

\noindent\textbf{View Sampling.}
We ablate the strategy to sample source and target views in Table~\ref{tab:view_sample}. Suppose there is only one source view, random sampling strategy randomly samples two views. One is source view and another is target view. Sampled views can be the same ones. In this case, VSA degrades to an autoencoder. Fixed sampling strategy firstly samples a random view as the source view. Second, it directly uses the next view as the target. If the source view is the last one, the first view will be used instead.

Table~\ref{tab:view_sample} shows that there is no difference between two strategies for fine-tuning. However, when doing linear probing, random-view strategy works better than fixed-view one. This is easy to understand since fixing sampling views makes the model easier to get overfitted to a local minimum. As a result, it does not learn enough knowledge for the task.

\begin{figure*}[t]
\centering
\includegraphics[width=1.0\linewidth]{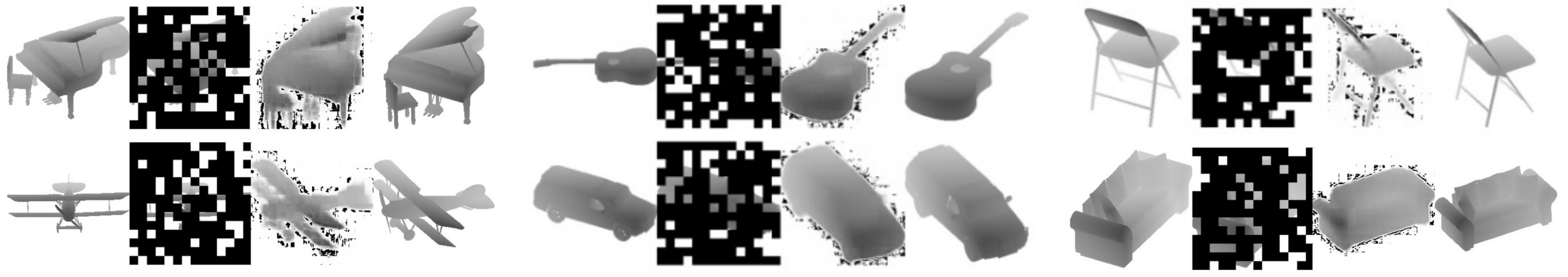}
\caption{\textbf{Example results of VSA+MAE on ModelNet40.} For each quadruplet, we show source image (first), masked image (second), image synthesized by VSA+MAE (third) and target image (forth).}
\label{fig:modelnet_mask}
\end{figure*}

\noindent\textbf{Cross Attention.}
As mentioned before, there are two kinds of blocks in VSA decoder, i.e., cross-attention and self-attention blocks. A cross-attention block can be deemed as a module fusing information from the target and source views. In Table~\ref{tab:cross_attention}, we ablate the number of cross-attention blocks in VSA decoder. Following the ViT-B decoder in MAE~\cite{he2021masked}, we set the number of blocks as 4. The number of cross-attention blocks ranges from 1 to 4. If there is no cross-attention block, the information from the source image is stuck, making no sense for synthesis.

In fine-tuning, increasing the number of cross-attention blocks helps in the beginning. However, if all blocks are changed into cross-attention ones, performance drops. It proves that self-attention blocks are also important in decoder.
Keys and queries of every cross-attention blocks are the same. They are not processed before the last block, making the latter not deep enough to extract their information.

However, as shown in Table~\ref{tab:cross_attention}, it is interesting to note that fine-tuning and linear probing are different in performance: using 1 block cross-attention is the best and 4 blocks do not hurt performance. Fine-tuning and linear probing have quite different properties. When evaluating the quality of pre-trained features, both pipelines should be considered.

\noindent\textbf{Number of Source Views.}
As mentioned before, there can be multiple source views in VSA. The features of multiple views are concatenated for cross-attention. We ablate the number of source views in Table~\ref{tab:num_views}. With 1, 2 and 4 input views, the fine-tuning performance drops. However, the linear probing performance is improved. It may be  because increasing input views improve synthesis, but sacrifice model potential for high-level tasks. When synthesis becomes easier, the model tends to learn in a low-level direction.

\noindent\textbf{Data Augmentation.}
We also ablate data augmentations for VSA in Table~\ref{tab:aug}. There are two kinds of augmentations in VSA: one for source image (src) and one for target image (tar). As shown in Table~\ref{tab:aug}, with minimal or no augmentation, VSA can work well. This advantage is similar to that of MAE~\cite{he2021masked}. 

Random crop on source image improves fine-tuning and linear probing performance. Color jitter on source image improves the linear probing. Based on VSA, adding data augmentation to source image helps the model to learn. However, random crop on target image can hurt both fine-tuning and linear probing performance. It is because synthesizing a randomly cropped image is impossible. The target image cannot be modified in the whole training process.

\begin{table*}[t]
\begin{center}
\begin{tabular}{llllll}
\hline\noalign{\smallskip}
 & &  & & \multicolumn{2}{c}{Classification Accuracy} \\
\multicolumn{1}{c}{Method}  & Backbone     & Data & View Select & (\textbf{Per-Class})   & (\textbf{Overall}) \\
\noalign{\smallskip}
\hline
\noalign{\smallskip}
VoxNet \cite{maturana2015voxnet}  & CNNs   & Voxels      & -           & 83.0 & 85.9      \\
PointNet \cite{qi2017pointnet} & CNNs &  Points        & -      &       86.2 & 89.2      \\
PointNet++ \cite{qi2017pointnet++} & CNNs   & Points     & -       & - & 91.9      \\
PointCNN \cite{li2018pointcnn} & CNNs  & Points    & -     &   88.1   & 91.8      \\
PTransformer~\cite{zhao2021point} & Transformer  &  Points & - & \textbf{90.6} & \textbf{93.7}  \\
\hline
GVCNN~\cite{feng2018gvcnn} & CNNs   & 12 Views     & yes          & 90.7 & 93.1 \\
ViewGCN~\cite{wei2020view} & CNNs  & 20 Views & yes  & \textbf{96.5} & \textbf{97.6} \\ 
ViewGCN \cite{wei2020view}$^*$ & CNNs & 12 Views & yes &    90.7   & 93.0 \\
MVTN~\cite{hamdi2021mvtn} & CNNs & 12 Views   & yes   & 92.0 & 93.8 \\
\hline
MVCNN~\cite{su2015multi} & CNNs     & 12 Views       & no        & 90.1  & 90.1 \\
VSA (ours) & Transformer & 12 Views  & no & \textbf{91.1} & \textbf{92.3} \\
\hline
\end{tabular}
\caption{\textbf{Comparisons with 3D classification methods.} Most view-based methods do view selection. It is fair to compare VSA with MVCNN and VSA works better. $^*$ indicates using MVTN rendering technology.}\label{table:3d_class}
\end{center}
\end{table*}

\noindent\textbf{Masking Ratio.}
Our VSA can be combined with MAE~\cite{he2021masked}. The source view image can be masked out part of patches before feeding into VSA encoder. We follow the MAE's default masking strategy of random sampling. As shown in Fig.~\ref{fig:mask}, a relatively high masking ratio (75\%) helps VSA to perform better.
However, a smaller masking ratio (10\%) hurts performance. The example synthesized images are in Fig.~\ref{fig:modelnet_mask}. Although there are some visual artifacts in the synthesized image, our VSA indeed synthesizes the target directions and textures.

\noindent\textbf{Training Schedule.}
The default training schedule in ablation study is 200 epochs. The influence of training schedule is shown in Fig.~\ref{fig:epoch}. With the training schedule becoming longer, the performance of fine-tuning improves significantly. However, a longer training schedule does not benefit linear probing. In VSA, after a short training schedule, the time is taken to improve the model potential.

\subsection{Comparisons with Previous Methods}

\begin{figure}[t]
\centering
\includegraphics[width=0.9\linewidth]{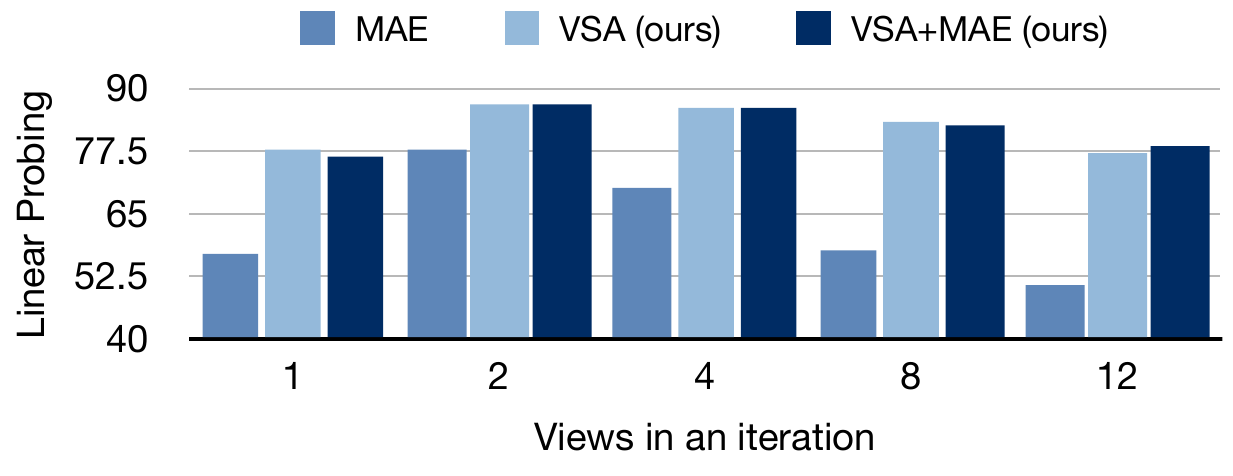}
\caption{\textbf{Comparisons with MAE.} VSA can work better with less input views in an iteration. VSA outperforms MAE for more than 20 points in linear probing.}
\label{fig:views}
\end{figure}

\begin{figure*}[t]
\centering
\includegraphics[width=1.0\linewidth]{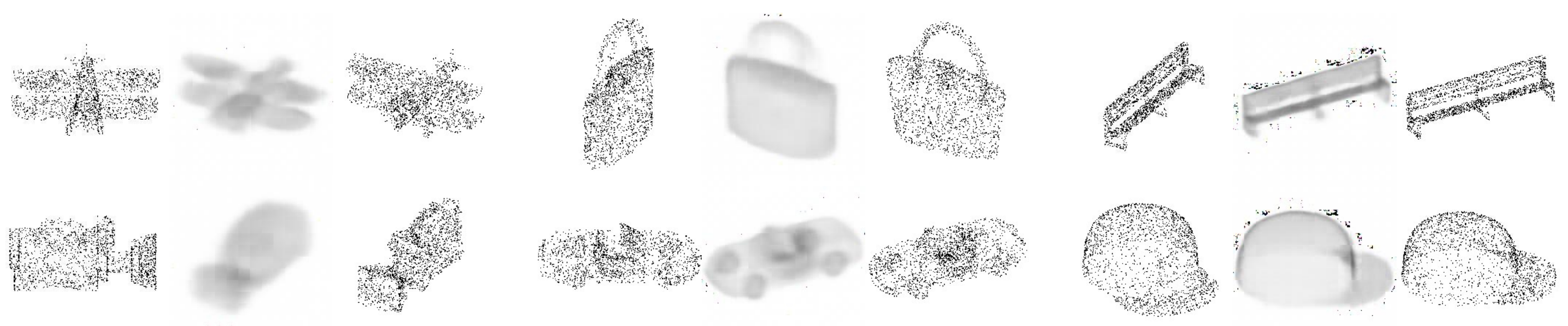}
\caption{\textbf{Results of VSA on ShapeNet Core55.} For each triplet, we show source image (left), image synthesized by VSA (middle) and target image (right).}
\label{fig:point}
\end{figure*}

\begin{table}[t]
\begin{center}
\begin{tabular}{llll}
\hline\noalign{\smallskip}
Dataset  & Scratch     & MAE & VSA (ours) \\
\noalign{\smallskip}
\hline
\noalign{\smallskip}
Shapenet Core55 & 67.8 & 79.0 & 79.9 \\
ScanObjectNN & 18.7 & 72.1 & 71.2 \\
\hline
\end{tabular}
\caption{3D classification accuracy on ShapeNet and ScanObjectNN (fine-tuning).}\label{table:point}
\end{center}
\end{table}

\noindent\textbf{Comparisons with MAE.}
First of all, we compare our VSA with recently proposed MAE~\cite{he2021masked}. We evaluate MAE and VSA on ModelNet40 using both end-to-end fine-tuning and linear probing. As shown in Fig.~\ref{fig:views}, VSA+MAE means using 75\% mask ratio for VSA source images. The default training schedule is 300 epochs.
The horizontal axis is the number of input views in an iteration.

In these two training pipelines, both MAE and VSA improve the scratch baseline a lot. For fine-tuning, there is a small difference between MAE and VSA when using all 12 views in each iteration. However, if the number of input views becomes smaller, VSA and VSA+MAE perform better than MAE. It proves that our VSA learns spatial-invariant features which can get more object information from fewer views. For linear probing, the difference between VSA and MAE is enlarged. VSA outperforms MAE for more than 20 points. In addition, in linear probing, more input views in an iteration may not bring better results, shown in Fig.~\ref{fig:views}.

\noindent\textbf{Comparisons on 3D Classification.}
We compare our VSA with other 3D classification methods in Table~\ref{table:3d_class}. Most of these methods use CNNs, while VSA is designed for vision transformer pre-training. Point transformer~\cite{zhao2021point} is a transformer designed for point-cloud data. VoxNet~\cite{maturana2015voxnet} uses voxels data. ViewGCN~\cite{wei2020view} uses high-quality rendering and 20 views. The performance of ViewGCN on MVTN~\cite{su2015multi}'s setting is also reported. Other view-based methods use 12 views -- most of them perform view selection. VSA uses common circular 12 views like MVCNN~\cite{su2015multi}. It is fair to compare VSA with MVCNN and VSA works better. We also find that our VSA outperforms GVCNN and ViewGCN in per-class accuracy. It means the spatial invariant features can make up for some shortcomings of a worse view selection strategy.

\subsection{Results on More Datasets}

\noindent\textbf{Results on ShapeNet Core55 and ScanObjectNN.}
We also do experiments on ShapeNet Core55 and ScanObjectNN. For these datasets, we render the point cloud to generate 2D images, resulting in density map 2D images. According to Fig.~\ref{fig:point}, VSA is able to synthesize this kind of images.
The transformer cannot directly generate these density maps. Instead, shaded images are synthesized. It means our VSA is trying to fit a continuous distribution, instead of discrete data points.

3D classification results are shown in Table~\ref{table:point}. The performance of VSA and MAE is comparable on this task. Both of them help the transformer to achieve good initialization. Combining them does not work for this data.

\noindent\textbf{View-synthesis on ScanNet.}
Our VSA also synthesizes views on ScanNet\cite{dai2017scannet}, as shown in Fig.~\ref{fig:scannet}. It proves that the potential of our VSA is not limited to synthetic data. In Fig.~\ref{fig:scannet}, the model is able to synthesize the door or chair that only has a small part in the source image. VSA can synthesize views for complex indoor scenes. It is more than geometry reconstruction. The synthesis process helps the model learn some high-level knowledge.

\section{Conclusions}

In this paper, we explore the situation when multi-view images are available. We propose view-synthesis autoencoders (VSA), a baseline for pre-training vision transformers with multi-view images. To be specific, VSA decoder uses source images features as value, source pose as key, target pose as query, to synthesize the target image by cross-attention. It is interesting to find that a simple transformer can achieve view synthesis. It also proves that spatial invariant features can be learned from this synthesis process. These features can benefit downstream tasks such as 3D classification on ModelNet40, ShapeNet Core55, and ScanObjectNN. VSA outperforms previous methods significantly for linear probing and is competitive for fine-tuning.

\clearpage
{\small
\bibliographystyle{ieee_fullname}
\bibliography{egbib}
}

\clearpage

\appendix

\title{Self-supervised Learning by View Synthesis}

\begin{table}
\centering
\begin{tabular}{ll}
config & value \\
\shline
optimizer & AdamW \\
base learning rate & 1.5e-4 \\
weight decay & 0.05 \\
optimizer momentum & $\beta_1, \beta_2{=}0.9, 0.95$ \\
batch size & 160 \\
learning rate schedule & cosine decay \\
warmup epochs & 40 \\
\shline
\end{tabular}
\caption{\textbf{Pre-training setting}}
\label{tab:impl_pretrain}
\end{table}
\begin{table}
\centering
\begin{tabular}{ll}
config & value \\
\shline
optimizer & AdamW \\
base learning rate & 1e-3 \\
weight decay & 0.05 \\
optimizer momentum & $\beta_1, \beta_2{=}0.9, 0.999$ \\
batch size & 160 \\
learning rate schedule & cosine decay \\
warmup epochs & 5 \\
training epochs & 100 \\
augmentation & Randaugment~\cite{cubuk2020randaugment} \\
\shline
\end{tabular}
\caption{\textbf{End-to-end fine-tuning setting}}
\label{tab:impl_finetune}
\end{table}
\begin{table}[h!]
\centering
\begin{tabular}{ll}
config & value \\
\shline
optimizer & AdamW \\
base learning rate & 1e-3 \\
weight decay & 0.05 \\
optimizer momentum & $\beta_1, \beta_2{=}0.9, 0.999$ \\
batch size & 320 \\
learning rate schedule & cosine decay \\
warmup epochs & 5 \\
training epochs & 100 \\
augmentation & Randaugment \\
\shline
\end{tabular}
\caption{\textbf{Linear probing setting}}
\label{tab:impl_linear}
\end{table}

\section{Network Architecture}

\begin{figure}
\centering
\includegraphics[width=1\linewidth]{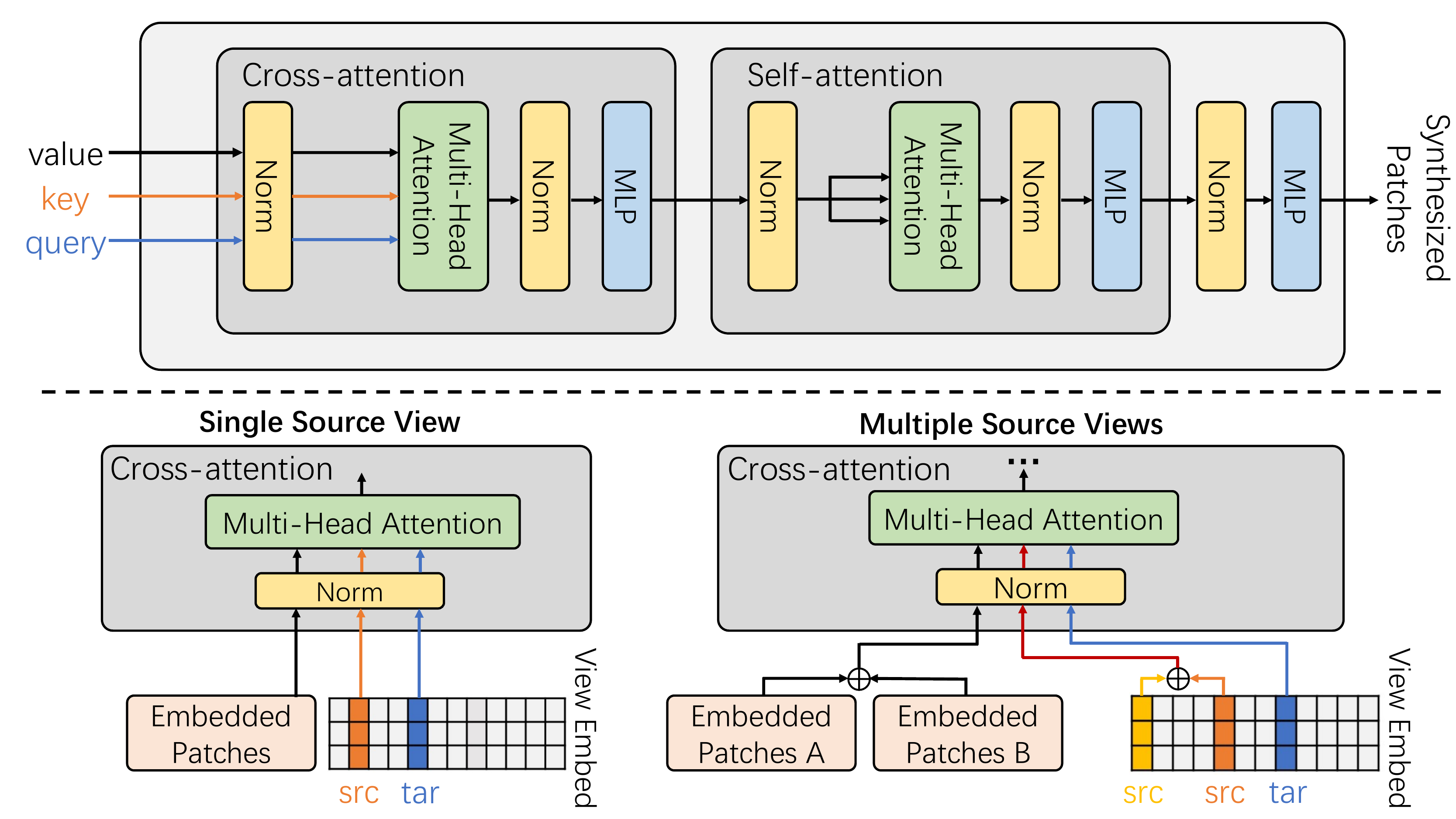}
\caption{\textbf{Architecture of VSA decoder.} Detailed architecture of VSA
decoder, which contains two cross-attention blocks and two self-attention blocks. Two cross-attention blocks share the same key and query}
\label{fig:arch_sup}
\end{figure}

Fig.~\ref{fig:arch} (upper) shows the detailed architecture of VSA decoder, supposing there are one cross-attention block and one self-attention block. Fig.~\ref{fig:arch_sup} shows the situation when VSA decoder has two cross-attention blocks and two self-attention blocks.

According to Fig.~\ref{fig:arch_sup}, two cross-attention blocks share the same key (orange arrow) and query (blue arrow). The value of first cross-attention block is the output of VSA encoder (black arrow). The value of second cross-attention block is the output of the first cross-attention block (black arrow).

\section{Training Details}

Training details are shown in Table~\ref{tab:impl_pretrain}, Table~\ref{tab:impl_finetune} and Table~\ref{tab:impl_linear}. The pre-training setting is in Table~\ref{tab:impl_pretrain}. Most of them are the same as setting in MAE~\cite{he2021masked}. Our VSA can work with a smaller batch size, like 160. It means our method can be used easily with 4 RTX 2080 Ti. In Table~\ref{tab:impl_finetune}, we can find fine-tuning setting is also similar to MAE. The only difference is that we use a smaller batch size.
The linear probing setting is in Table~\ref{tab:impl_linear}. We find there is little difference between LARS~\cite{you2017large} and AdamW when using a small batch size. To simplify our setting, we keep using AdamW for linear probing. Besides, although MAE changes weight decay, learning rate and augmentation for linear probing, we find their influence is small for VSA. As a result, we use the same learning rate, weight decay and augmentation as fine-tuning. We use 320 as the batch size for linear probing. In conclusion, our VSA can work without carefully selected parameters. Most parameters in MAE can be used in VSA.

\begin{figure*}
\centering
\includegraphics[width=0.9\linewidth]{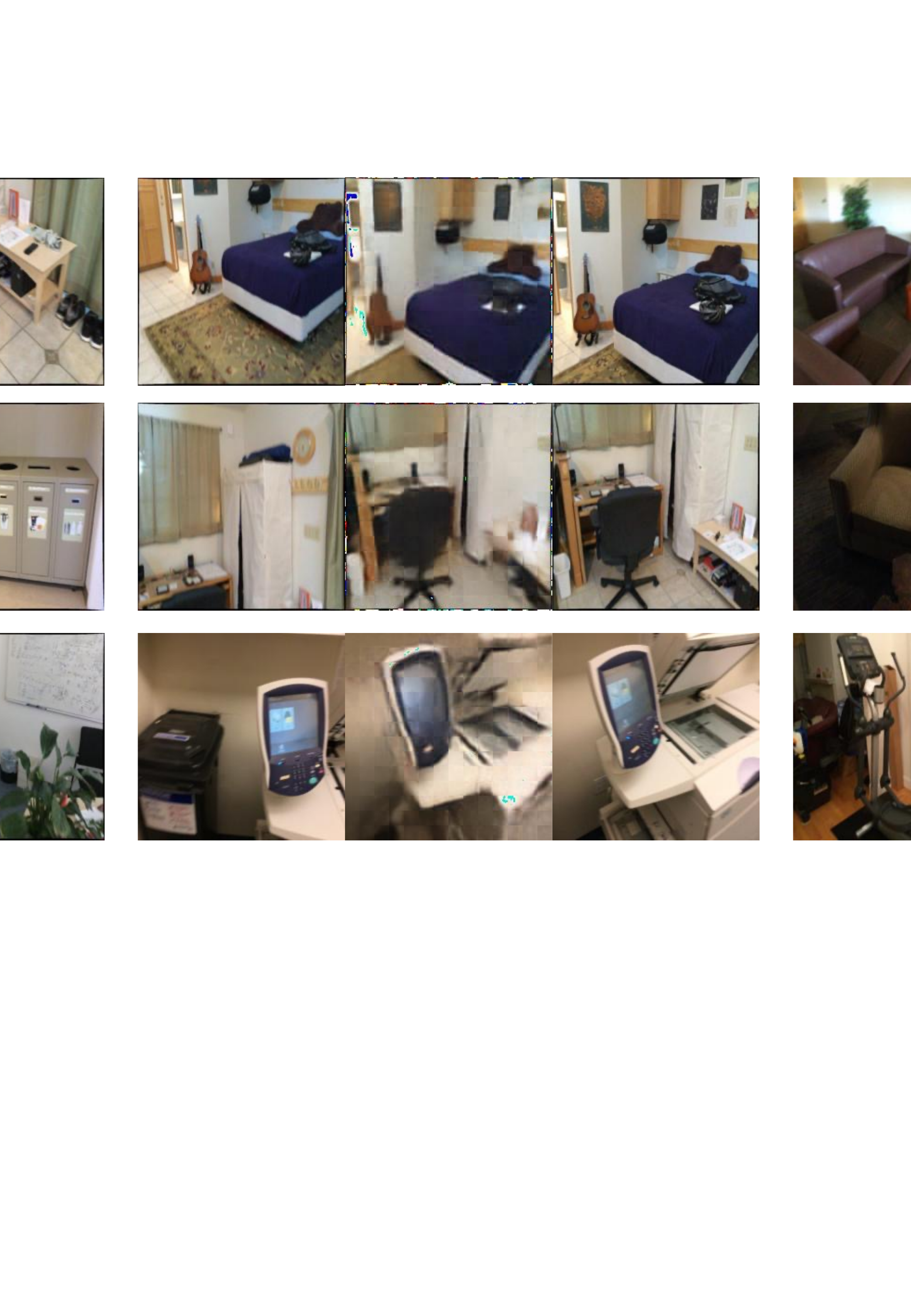}
\caption{Additional results on Scannet~\cite{dai2017scannet}. For each triplet, we show source image (left), image synthesized by VSA (middle) and target image (right)}
\label{fig:scannet_sup}
\end{figure*}

\end{document}